\documentclass[10pt,twocolumn,letterpaper]{article}

\usepackage[pagenumbers]{cvpr} 

\usepackage{graphicx}
\usepackage{amsmath}
\usepackage{amssymb}
\usepackage{booktabs}

\usepackage[pagebackref,breaklinks,colorlinks]{hyperref}

\usepackage[capitalize]{cleveref}
\crefname{section}{Sec.}{Secs.}
\Crefname{section}{Section}{Sections}
\Crefname{table}{Table}{Tables}
\crefname{table}{Tab.}{Tabs.}

\def\cvprPaperID{*****} 
\def\confName{CVPR}
\def\confYear{2022}

\begin{document}

\title{Situation Awareness for Automated Surgical Check-listing in AI-Assisted Operating Room}

\author{Tochukwu Onyeogulu$^{1}$, Salman Khan$^{1}$, Izzeddin Teeti$^{1}$, Amirul Islam$^{1}$, Kaizhe Jin$^{2}$, 
\\
{Adrian Rubio-Solis$^{2}$, Ravi Naik$^{2}$, George Mylonas$^{2}$, Fabio Cuzzolin$^{1}$}
\\
$^{1}$ Visual Artificial Intelligence Lab (VAIL), Oxford Brookes University
\\
$^{2}$ Imperial College London
\\
{\tt\small $^{1}$ 19175136, salmankhan, iteeti, aislam,  fabio.cuzzolin@brookes.ac.uk}\\
{\tt\small $^{2}$ k.jin20, arubioso, ravi.naik15, george.mylonas@imperial.ac.uk}
}

\maketitle

\begin{abstract}
Currently, more surgical procedures are being performed using minimally invasive surgery (MIS). This is because of its many benefits, such as minimal postoperative complications, less bleeding, minor scarring, and rapid recovery. However, MIS's constrained field of view, small operating room, and indirect viewing of the operating scene could lead to surgical tools colliding and potentially harming human organs or tissues. Therefore, MIS problems can be considerably reduced, and surgical procedure accuracy and success rates can be increased using endoscopic video feed to detect and monitor surgical instruments in real-time. In this paper, a set of improvements made to the YOLOV5 object detector to enhance the detection of surgical instruments was investigated, analyzed, and evaluated. In doing this, we performed performance-based ablation studies, explored the impact of altering the YOLOv5 model's backbone, neck, and anchor structural elements, and annotated a unique endoscope dataset. In addition, we compared the effectiveness of our ablation investigations with that of four additional SOTA object detectors (YOLOv7, YOLOR, Scaled-YOLOv4, and YOLOv3-SPP). With the exception of YOLOv3-SPP, which had the same model performance of 98.3 percent in mAP and a similar inference speed, all our benchmark models, including the original YOLOv5, were surpassed by our top refined model in experiments using our fresh endoscope dataset.

\end{abstract}

\section{Introduction}
\label{sec:intro}
According to a report by the World Health Organisation (WHO), 25\% of surgical procedures performed out on patients worldwide lead to post-surgical complications \cite{ref1}. Seven million people in this population have serious post-surgical complications, of which 14\% of these patients die as a result of these post-surgical complications. Previous studies have shown that over 60\% of these complications are preventable, which implies that an improvement in surgical procedures through the integration of advanced AI solutions can reduce post-surgical complications \cite{ref2,ref3}. Earlier studies by \cite{ref4,ref5} suggest that 15.9\% of the errors encountered during surgical operations are related to surgical instruments. These errors result from errors made by surgeons or assistant surgeons in identifying the correct surgical instruments, poor decision-making, communication failure, impaired vision, distractions, workload, and cognitive burden. In response, WHO proposed the use of surgical safety checklists (SSC) to reduce errors in surgical operations. Although these checklists have been proven to reduce the death rate by 0.7\% and patient complications by 3\%, they rely heavily on manual data entry and assessment, which are subjective to the observers' bias. Calls have since been made to fully exploit the integration of data from multiple sensors to mine the streams of information that characterise a surgical procedure and to provide early warnings to surgeons of deterioration in cognitive performance. The MAESTRO Jr. The AI concept aims to achieve this goal by laying the foundations for the mid-21st century operating room. A surgical environment powered by a trustworthy, human-understanding artificial intelligence system is able to continually adapt and learn the best way to optimise safety, efficacy, teamwork, economy, and clinical outcomes. To achieve this aim, MAESTRO objectives are (i) to deploy and test a platform integrating multiple sensing modalities as a sandpit for further research; (ii) to validate the multi-sensor fusion sensing of escalating cognitive load; (iii) to perform multi-modal situation awareness for automated surgical check-listing; and (iv) to study the feasibility of continual learning for adapting to new surgical teams and procedures, as the pillar of an AI-assisted operating room in the future.

\section{Detection From Laparoscopic Video}

\subsection{Data Collection}
The endoscope dataset consists of two videos of a surgeon performing laparoscopic cholecystectomy procedures on a patient using a standardized porcine model. Cholecystectomy, also known as gallbladder removal, is a tiny pouch-like organ located in the top right corner of the human stomach. It stores bile, a fluid released by the liver that assists in the digestion of fatty foods. Because humans do not require a gallbladder, surgical procedures to remove it are widely prescribed if a patient experiences any challenges. The endoscope videos were recorded using a high-definition camera (1080p, 1,920x1,080 resolution and progressive scan) Karl Storz, 30 degrees, 10mm laparoscope. It was connected to a Karl Storz Image 222010 20 SCB Image 1 Hub Camera Control Unit. Before beginning the required annotation process, video data were collected as part of MAESTRO Jr. The AI project at Imperial College London was anonymized. Our aim in this section is to describe the various procedures employed in the data collection process, which includes the surgical task and the surgical phases involved in the data collection process. We also discuss the data annotation tool used for annotating the dataset, data annotation protocols, and data processing steps involved in preprocessing the data to prepare it for training.
\subsubsection{Surgical tasks}
The various surgical tasks involved in the data collection process include 
\begin{itemize}
    \item Laparoscopic cholecystectomy using a standardised porcine model.
    \item Body torso laparoscopic box trainer with prepositioned trocars. Possibly a haptic box trainer.
    \item Basic laparoscopic stack and standard surgical equipment (hook diathermy, Maryland graspers, crocodile graspers, 10mm and 5mm trocars, 5/10mm Endoclip applicator, Bert bag/Endocatch, 30-degree camera).
\end{itemize}

\subsubsection{Surgical Phases}

This subsection provides a brief description of the various surgical phases involved in the data collection process.

\textbf{Retraction of the gallbladder:} The surgeon passes the instrument(s) to the assistant surgeon, who holds the flaps up, in order for the surgeon to have a better view of the regions of interest, which are the gallbladder, liver, and cystic duct. 

\textbf{Dissection of critical view of safety:} The surgeon uses hook diathermy and a Maryland grasper to dissect the hepatocystic/calot’s triangle.

\textbf{Clipping and division of the cystic duct:} The surgeon uses endoclip and laparoscopic scissors to clip the region of interest, which in this case is the cystic duct.

\textbf{Clipping and division of the cystic artery:} The surgeon uses endoclips and laparoscopic scissors to clip the region of interest, which in this case is the cystic artery. 

\textbf{Dissection of the gallbladder from the liver bed or cystic plate:} The surgeon uses a crocodile grasper and hook diathermy to dissect the gallbladder from the liver bed. This causes the presence of smoke in the surgical scene.   

\textbf{Removal of the gallbladder:} The surgeon uses a Bert bag or Endocatch to remove the gallbladder from the surgical scene. 

It is worth mentioning that the participants involved in the data collection process were high surgical trainees specializing in general surgery, recruited from the Department of Surgery and Cancer, Imperial College London. The participants had previous exposure to basic laparoscopic surgical training and simulated and clinical exposure to laparoscopic surgical/cholecystectomy training. 

\subsubsection{Data Annotation Protocol}
For the object detection task, the endoscope videos were manually annotated by drawing bounding boxes around the tips of the surgical instruments of interest in each image frame and entering the corresponding class label associated with each bounding box. This is what we term the "ground truth labels." For this annotation task, we annotated eight different surgical instruments: Crocodile grasper, Johan grasper, hook diathermy, Maryland grasper, clipper, scissors, bag holder, and trocar.

\subsubsection{Data Annotation Tool}

\begin{figure}[!t]
\centering
\includegraphics[width= 0.5 \textwidth]{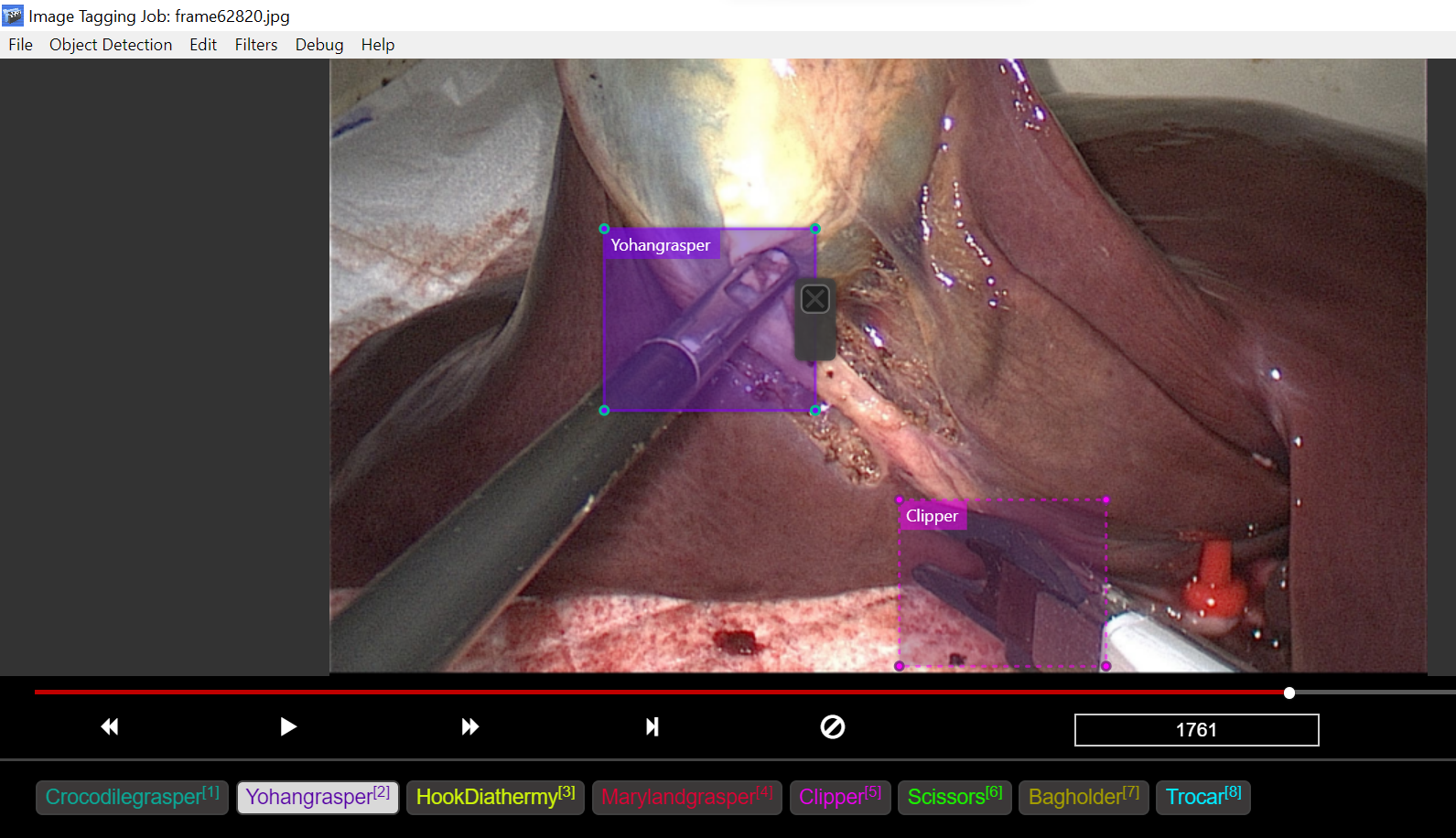}
\caption{\label{fig:vott} Screenshot of using Microsoft VoTT for annotating of our novel endoscope dataset.}
\end{figure}

\textbf{VoTT:} The virtual object tagging tool (VoTT), a free and open-source data annotation tool from Microsoft, was used to annotate our endoscope video dataset. VoTT was used to draw bounding boxes around the tips of the surgical instruments of interest in the dataset. Figure \ref{fig:vott} depicts a screenshot of its graphical user interface.

\begin{figure}[!t]
\centering
\includegraphics[width= 0.48 \textwidth]{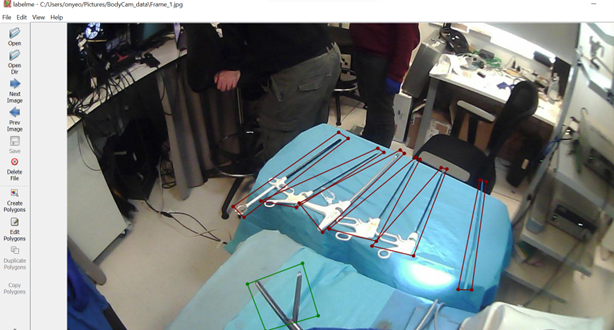}
\caption{\label{fig:labelme} Screenshot of using LabelMe for annotating of our novel endoscope dataset.}
\end{figure}

\textbf{LabelMe:} LabelMe is a free open-source data annotation tool developed by MIT for manual image annotation in object detection, classification, and segmentation tasks. LabelMe was created in Python with the goal of collecting a large collection of images with ground-truth labels. LabelMe is simple to use and allows polygons, circles, rectangles, lines, line strips, and points to be drawn, among other shapes. Fig. \ref{fig:labelme} provides an overview of the use of LabelMe for image annotation.

\subsubsection{Data Preprocessing}

To prepare our novel endoscope dataset for training, we performed a series of data preprocessing steps. We developed a Python script to convert our novel endoscopic videos into image frames. We implemented the Microsoft VoTT annotation tool to annotate the dataset and performed an adaptive image-scaling operation to obtain a standard image size of 640×640 for training. The Microsoft VoTT annotation tool provides the annotation of our dataset in the MSCOCO data format, which is not an acceptable data format for YOLO algorithms. A Python script was developed to convert the annotation from the MSCOCO data format to YOLO data format.

\subsection{Methodology}

YOLOv5, which is the fifth generation of the YOLO family, was released by Glenn Jocher and his research team at Ultralystics LLC a few months after the release of YOLOv4 by Bochkovskiy et al. \cite{ref27}. Earlier versions of YOLO models were developed using a custom Darknet framework, which was written in the C programming language. However, Glenn Jocher and his research team changed the trajectory by utilizing PyTorch, a deep learning library developed by the Facebook research team and written in the Python programming language, to build the YOLOv5 model. Small, medium, large, and Xlarge are the four distinct scales that YOLOv5 offers for their models. While the general structure of the model remains unchanged across different scales, the size and complexity of each model are adjusted by a different multiplier for each scale. Although all our modifications and experiments were carried out using the large YOLOv5 model, it can still be replicated across other variants of the YOLOv5 model by adjusting the width and depth multipliers.
To establish a baseline, we trained and evaluated the unaltered YOLOv5 model using our novel endoscopic dataset. Next, we implemented our proposed modifications ( discussed in the next section) on the YOLOv5 model, trained and evaluated its performance, and conducted an ablation study using our baseline unmodified YOLOv5 model. This procedure was repeated to monitor whether certain refinement strategies enhanced or compromised one another, while gradually adding and removing more complex combinations. Finally, we benchmarked the results from our unrefined and refined YOLOv5 models with those from YOLOv7, Scaled-YOLOv4, YOLOR, and YOLOv3-SPP models trained on our novel endoscope dataset. A comparison of their accuracy in detecting surgical instruments from endoscope videos and identification of the best possible detector for our MAESTRO endoscope vision-based system. It is worth mentioning at this point that all models trained on our novel endoscope dataset were trained from scratch for 300 epochs without using pre-trained weights.

\begin{figure*}[!t]
\centering
\includegraphics[width= 0.7 \textwidth]{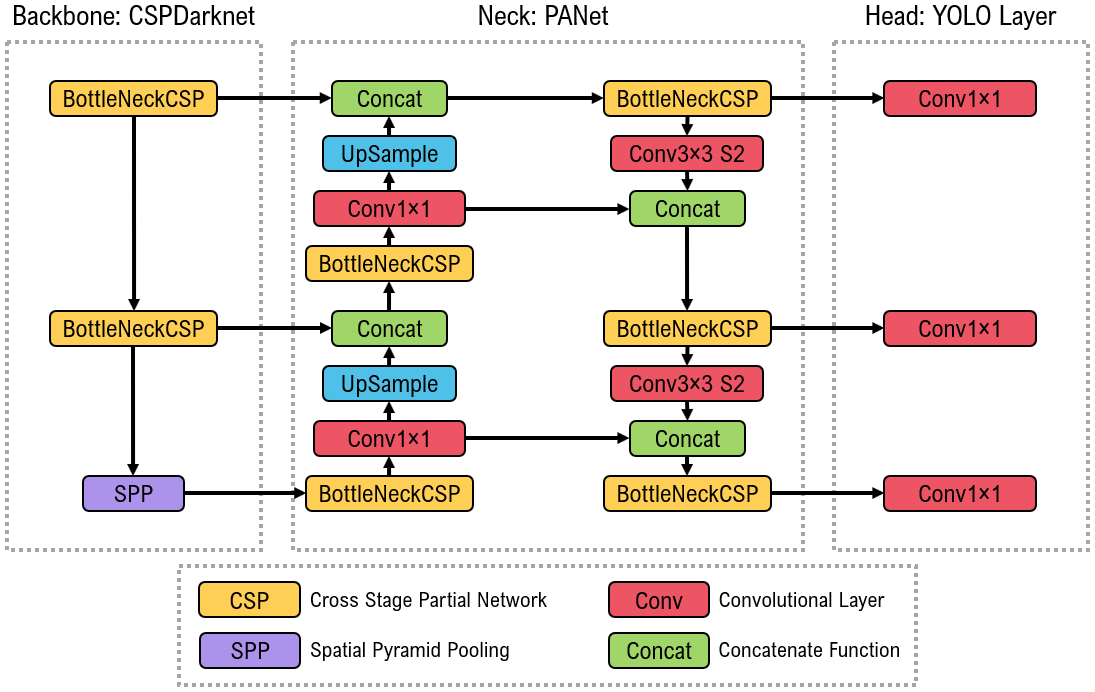}
\caption{\label{fig:yolov5}The unmodified YOLOv5 model's architecture, which is divided into three parts: Backbone:CSPDarknet with an SPP layer, Neck:PANet, and Head:YOLOLayer The data is first supplied into CSPDarknet for feature extraction before being loaded into PANet for feature fusion. Finally, the YOLO Layer generates the object detection results (i.e., class label, class score, location, size)\cite{ref87}.}
\end{figure*}

\subsection{Overview of Yolov5 Architecture}
\label{overview}

The YOLOv5 architecture\cite{ref87} illustrated in Fig. \ref{fig:yolov5},  employs the Cross Stage Partial Darknet53 (CSPDarknet53) architecture and spatial pyramid pooling (SPP) layer as its backbone, the path aggregation network (PANet) architecture as the neck, and the YOLOv3 head for detection. The CSPDarknet53 module primarily extracts rich information from input images by performing feature extraction on the feature map. The output of CSPDarknet53 is passed through the SPP layer before being sent to the neck for feature aggregation. The SPP layer, which is placed between the backbone and neck, extracts important features from several scales into a single feature map, which increases the detection performance. In the neck, the PANet architecture, which is an extension of the FPN with an additional bottom-up path, improves the ability of the model to detect objects at different scales by aggregating strong semantic feature maps from different feature layers. The head, which is the last step in the detection process, makes dense predictions on multiscale feature maps from the neck module. The dense prediction consists of bounding box coordinates (center, height, and width), class labels, and confidence scores. To adapt to the differences in datasets, YOLOv5 incorporates the use of adaptive anchor-box computation on its input. This allows the YOLOV5 model to automatically learn the best anchor box for any given dataset and to utilize it throughout the training process \cite{ref82}.

\subsection{Our Modifications}
The YOLOv5 model makes use of a .yaml file which contains instructions for the overall model architecture. These instructions are fed to the parser, which then builds the model based on the information in the .yaml file. In order for us to implement any modifications, we emulated this arrangement by rewriting a new .yaml file containing our proposed modification to instruct the parser in building the model.
The key modifications that we proposed for the YOLOv5 models are the backbone, neck, and a few hyper-parameter changes to optimize the model performance on our novel endoscope dataset. This modification was inspired because the original YOLOv5 model was trained on the MSCOCO dataset, which is different from our novel endoscope dataset. In this section, we describe our proposed modification of the YOLO5 network in detail.

\subsubsection{Backbone} 
The unmodified YOLOv5 model uses CSPDarknet as its backbone. In this paper, we replaced the CSPDarknet backbone with a modified VGG-11 backbone by removing the fully connected layers in the original VGG-11 architecture and adding the SPP layer, which is an additional layer between the unmodified YOLOv5 backbone and neck. VGG, which represents the virtual geometry group, was first proposed by A. Zisserman and K. Simonyan at the University of Oxford. They investigated the influence of convolutional neural network depth on accuracy in the context of large-scale image recognition \cite{ref64}. Their main contribution was to increase the depth of the model by replacing the large kernel-size filters with very small 3 × 3 convolution filters. This achieved significant improvement over the prior art, such as AlexNet \cite{ref83} and almost 92.7\% top-5 test accuracy on the ImageNet dataset.

\begin{figure}
     \centering
     \begin{subfigure}[b]{0.15\textwidth}
         \centering
         \includegraphics[width=\textwidth]{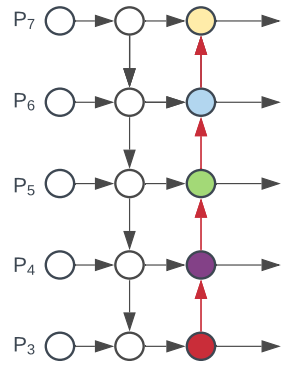}
     \end{subfigure}
     \begin{subfigure}[b]{0.15\textwidth}
         \centering
         \includegraphics[width=\textwidth]{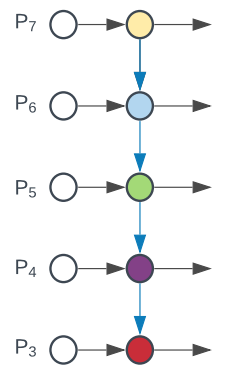}
     \end{subfigure}
     \begin{subfigure}[b]{0.15\textwidth}
         \centering
         \includegraphics[width=\textwidth]{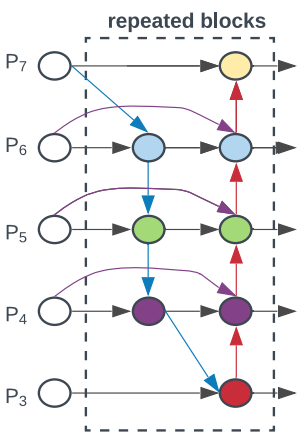}
     \end{subfigure}
        \caption{Overview of the three different neck architectures we experimented with: Path aggregation network (PANet), Feature Pyramid Network (FPN) and Bi-Feature Pyramid Network (Bi-FPN) (Left-Right) \cite{ref69}.}
        \label{fig:neck}
\end{figure}

\subsubsection{Neck}

The neck in YOLOv5 is a series of layers between the backbone and the head. The unmodified YOLOv5 model uses a path aggregation network(PANet) as its neck architecture. In this paper, we experimented with two other neck architectures: a feature pyramid network (FPN) and a bi-directional feature pyramid network (Bi-FPN). The architecture is shown in Fig. \ref{fig:neck}.

\textbf{Feature Pyramid Network (FPN)} is not an object detector. This is a feature extractor architecture incorporated within an object detector. The feature pyramid network extracts scaled feature maps at multiple layers from a single-scale input image of any size and passes these extracted feature maps to the head of the object detector (for example, YOLOv3 head), which performs the detection task. This process is not influenced by the backbone of the object detector. A feature pyramid network is composed of single bottom-up and top-down information pathways. The bottom-up information pathway is the backbone of the object detector, which generates feature maps of varying sizes using a size-two scaling step. The top-down information pathway uses upsampling techniques (with two nearest neighbors) on the previous layer, which is then linked with feature maps from the final layer of each stage in the bottom-up information pathway using a skip connection. Each connection between feature maps from the bottom-up information pathway to the top-down information pathway has the same spatial size \cite{ref68}.  

\textbf{Bi-Feature Pyramid Network (BiFPN)} also known as Weighted Bi-directional Feature Pyramid Network, is an improved Feature Pyramid Network (FPN) developed by the Google Research Brain Team \cite{ref69}. BiFPN integrates the notion of multi-level feature fusion from feature pyramid networks (FPN), path aggregation networks (PANet), and neural architecture search-feature pyramid networks (NAS-FPN). Hence, it enables simple and rapid multi-scale feature integration. Major modifications integrated into the bidirectional feature pyramid network are (i) removal of nodes with only one input. Nodes with one input edge and no feature integration contribute less to feature networks seeking to fuse distinct features. (ii) Additional edges from the original input node to the output node of the same grade to integrate more features without increasing the computational complexity \cite{ref69}.

\subsection{Benckmark Object Detection Model}

In order to justify, measure, and validate the performance of our modified and unmodified YOLOv5 models on our novel endoscope dataset, we benchmarked our model performance results with those of popular object detectors trained on our novel endoscope dataset. This section provides a brief review of some selected object detection models. We benchmark our model with and justify why each object detector was chosen as part of our benchmark models.  
\subsubsection{YOLOv3-SPP}
Although the YOLOv3 \cite{ref26} object detection model achieved state-of-the-art performance with respect to its speed and accuracy, it still had a short fall with scale variation, which needed improvements since its multi-scale object detection capabilities are linked to its network's receptive fields. YOLOv3-SPP \cite{ref77} was proposed to ameliorate this problem by introducing a spatial pyramid pooling layer into the YOLOv3 network to efficiently tackle the scale variation problem and integrate multi-scale features effectively. This equips the network with the ability to learn the various objects' features properly. This approach achieved an improved object detection performance (mAP) over YOLOv3 when experimented on the UA-DETRAC dataset. which is our justification for selecting it as one of our benchmark models.

\subsubsection{Scaled-YOLOv4}

The Scaled-YOLOv4 object detection model designed by Chien-Yao Wang, Alexey Bochkovskiy, and Hong-Yuan Mark \cite{ref75} propelled the YOLOv4 model forward by introducing a network scaling strategy which doesn't only scale the depth, breadth, and resolution of the network but also scales the structure of the network. Their strategy outperformed the previous state-of-the-art object detection model, EfficientDet \cite{ref69}, developed by the Google Brain Research team, on both ends of the speed versus accuracy frontier when tested on the MSCOCO \cite{ref85} dataset.

\begin{figure*}[!t]
\centering
\includegraphics[width= 0.7 \textwidth]{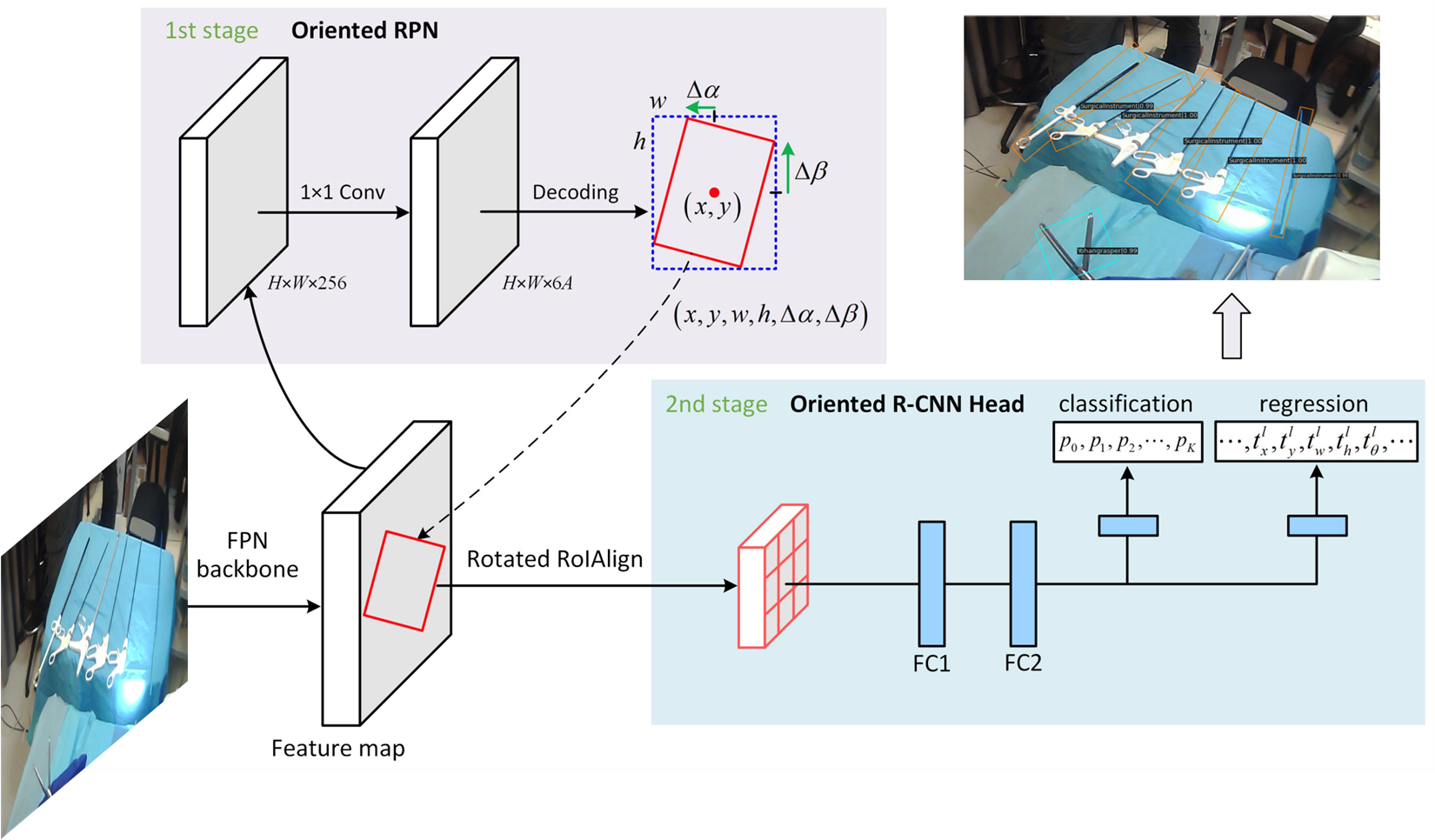}
\caption{\label{fig:rcnn}Overall framework of oriented R-CNN \cite{xie2021oriented} consists of two stages. The first stage generates oriented proposals by oriented RPN and the second stage is oriented R-CNN head to classify proposals and refine their spatial locations.}
\end{figure*}

\subsubsection{YOLOR}

YOLOR \cite{ref84} which stands for “You Only Learn One Representation", is a state-of-the-art object detection model which is a different variant in the YOLO series (i.e. YOLOv1 to YOLOV7) because it utilises a unified framework to concurrently encode implicit and explicit knowledge. This unified network can produce a unified representation to fulfil several jobs at the same time. The idea behind the YOLOR framework was adopted from humans' ability to implicitly (learning subconsciously) or explicitly (through regular learning) comprehend their surroundings through perception, listening, touch, and memory. The authors also claimed that YOLOR can perform kernel space alignment, prediction refinement, and multi-task learning in a Convolutional neural network. Results from using the YOLOR network suggest that the inclusion of implicit knowledge into the network model improves the performance of the network on various tasks when experimented on the MSCOCO \cite{ref85} dataset, YOLOR achieved comparable accuracy with Scale-YOLOv4 \cite{ref75} but an inference speed 88\% faster than Scale-YOLOv4.

\subsubsection{YOLOv7}

YOLOv7 \cite{ref29} which is the most recent object detection model within the YOLO family, was developed by Alexey Bochkovskiy, who took up the management of the YOLO algorithm after the original author, Joseph Redmon, stopped computer vision research due to ethical issues, and Kin-Yiu, Wong who joined the computer vision research domain with the innovation of cross-spatial networks, allowing YOLOv4 and YOLOv5 to construct more scalable backbone networks. YOLOv7 achieved state-of-the-art performance on the MSCOCO dataset \cite{ref85} when compared with its peers. In order to achieve this height, the authors' major contributions to the YOLOv7 model are: (i) introduction of an extended version of the efficient layer aggregation (ELAN) computational block, which they termed E-ELAN as their final aggregation layer. (ii) introduction of a novel model scaling technique, which can scale both the network depth and breadth simultaneously while concatenating layers together. (iii) addition of an auxiliary head network that is used to supervise the detection head during training and a model re-parameterization technique in order to make the model more robust and generalize well on new data. 

\begin{figure}[!t]
    \centering
    \includegraphics[width= 0.45 \textwidth]{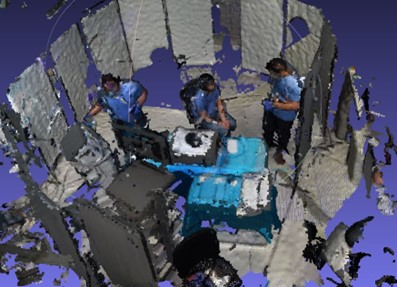}
    \caption{Sample of Pointcloud.}
    \label{fig:pc_sample}
\end{figure}

\subsection{Model Evaluation}

\begin{figure*}[!t]
    \centering
    \includegraphics[width= 0.95 \textwidth]{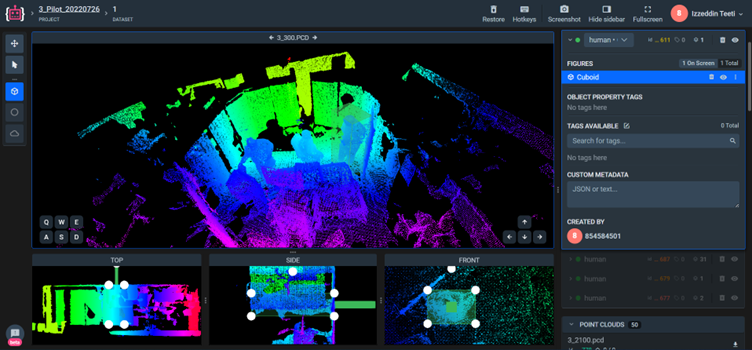}
    \caption{Interface of Supervisely to annotate Pointclouds.}
    \label{fig:supervisely}
\end{figure*}

The evaluation metrics adopted in evaluating the performance of our model are F1 score, mAP, and mAP@0.5:0.95. The mAP is chosen as our primary evaluation metric, followed by the F1-score. For each class label in our dataset, we first calculated the precision and recall rate using true positives (TP), false positives (FP), and false negatives (FN), as described in \eqref{eq1} and \eqref{eq2}. Using the results obtained from computing the precision and recall, we obtained the F1-score (see Equation \eqref{eq3}). The F1-score, which is the harmonic mean of precision and recall, has a maximum possible value of 1, suggesting excellent precision and recall scores, and a minimum possible value of 0, suggesting a poor recall or precision score. The mAP is computed by averaging the average precision of all class labels in our dataset(see equation \eqref{eq4}). In this paper, mAP with an IoU threshold of 0.5 was used as our evaluation criterion for measuring the performance of our model.
\begin{align} 
\text{Precision} = \frac{TP}{TP + FN}
\label{eq1} \\
\text{Recall}  = \frac{TP}{TP + FP}
\label{eq2} \\
\text{F1 Score}  = 2 *\frac{\text{Precision} * \text{Recall}}{\text{Precision} + \text{Recall}} \label{eq3}\\
\text{mAP}   = \sum_{n=1}^{N} \frac{\text{Average Precision} (n)}{N}
\label{eq4}
\end{align}

\section{Rotated Object Detection Methods}

\subsection*{Oriented R-CNN}
For the detection of tools using the bodycam, we fine-tuned an existing rotated object detector called oriented Region-based Convolutional Neural (R-CNN) \cite{xie2021oriented}, which consists of an oriented RPN and an oriented R-CNN head (see Fig. \ref{fig:rcnn}). It is a two-stage detector, where the first stage generates high-quality oriented proposals in a nearly cost-free manner and the second stage is oriented R-CNN head for proposal classification and regression.

\section{Surgical Phase Segmentation}
\label{sec:phase_seg}
\subsection{Methodology}
To locate the action (phase) temporally and spatially, the surgical tool in the scene that performs the action must be detected, and its temporal and spatial dimensions should be modelled. To this end, our methodology consists of three main phases. First, a detection method is applied to detect the surgical tools in each frame. The surgical tools were tracked through frames to form tubes. Then, the surgical tool features are extracted from the tubes using a 3D feature extraction method to connect these features locally (in the same frame) and globally (across frames) using local and global graphs, respectively. We have three types of local graphs, that is, \textbf{ fully connected }, where each of the surgical tool tubes is connected to all the other tubes, \textbf{scene} representing the tree-like structure where all the surgical tool tubes are connected to the scene only, and \textbf{scene with the same label} having the same properties as the scene with the connection between the same surgical tools.  This method allows surgical tools to attend to each other in the same frame (in the space domain), which will lead to locating the action and across frames (in the time domain), leading to the detection of the start and end of the action.  

\section{3D Detection from Pointcloud}
\label{sec:3d_det}
One of MAESTRO’s tasks is to be able to detect medical staff and tools in 3D space, which provides a better spatial understanding inside the operating room. This document aims to explain the relevant aspects of 3D detection in the MAESTRO project. 

\subsection{Data Preparation }

\subsubsection{Data Collection}
The dataset consists of Pointclouds that were extracted after calibrating and combining eight Azure Kinect RGB-D cameras, each having a 1 MP time-of-flight (Microsoft Team, n.d.). We used eight cameras to cover all angles of the operating scene. The collected dataset consists of four sessions, each has 4000 Pointclouds. For the first training, we used 150 Pointclouds. We present one of the Pointcloud samples shown in Fig. \ref{fig:pc_sample}.

\begin{figure}[!t]
     \centering
     \begin{subfigure}[b]{0.4\textwidth}
         \centering
         \includegraphics[width=\textwidth]{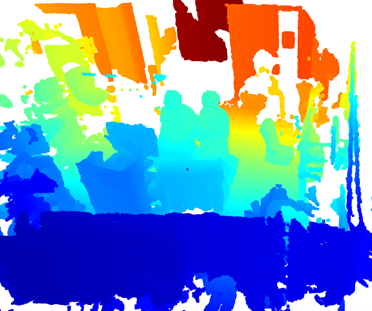}
         \caption{}
     \end{subfigure}\\
     \begin{subfigure}[b]{0.4\textwidth}
         \centering
         \includegraphics[width=\textwidth]{Figures/pointcloud_bin.png}
         \caption{}
     \end{subfigure}
     \caption{Pointcloud in (a) PCD and (b) bin format.}
     \label{fig:pc_format}
\end{figure}

\subsubsection{Data Annotation}
We manually annotated the Pointclouds by drawing a 3D bounding box. The objects of interest were humans with the class name ‘Human’. We used an online annotation platform called Supervisely (shown in Fig. \ref{fig:supervisely}) (Supervisely: Unified OS for Computer Vision, n.d.), which has the advantage of copying bounding boxes across frames which, in turn, allowed faster-annotating procedure. Furthermore, it has a taskforce environment that splits the annotation work among annotators to supervise and validate the annotations by reviewers or administrators.

\subsubsection{Data Preprocessing}
The data are preprocessing following the steps as listed below:
\begin{itemize}
    \item [1.] We downsampled the combined Pointclouds to end up with Pointclouds of lower size, which will also take less time to process than the original Ponitclouds.
    \item [2.] Then we converted the Pointclouds from .ply format to .pcd format which is the form that Supervisley accepts. 
    \item [3.] After annotating the Pointcloud, we downloaded them along with their annotations. Then both the Pointclouds and their annotations were converted from .pcd and .json formats to .bin and .txt formats, respectively. This is because most of the 3D detection models were trained and tested on Autonomous Driving datasets like KITTI \cite{geiger2012we}, nuScenes \cite{caesar2020nuscenes}, and Waymo(Addanki et al., 2021), which have their Pointclouds in .bin format. We provide the samples of Pointclouds for pcd and bin format in Fig. \ref{fig:pc_format}(a) and \ref{fig:pc_format}(b), respectively.
\end{itemize}

\subsection{Methodology}
\subsubsection{PV-RCNN}

\begin{figure*}[!t]
    \centering
    \includegraphics[width= \textwidth]{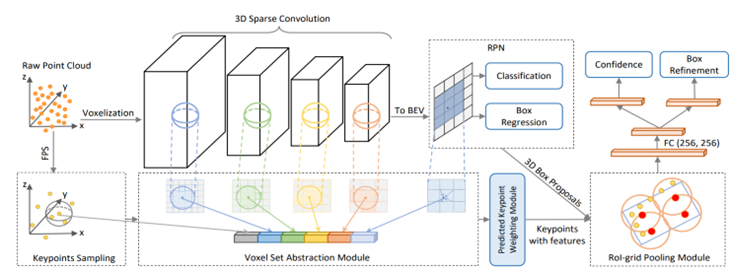}
    \caption{PV-RCNN Structure \cite{Shi2020}.}
    \label{fig:pv_rcnn}
\end{figure*}

We used an off-the-shelf 3D detector called Point-Voxel-RCNN (PV-RCNN) \cite{Shi2020}. 3D detection models can be categorised into two main categories, Point-based and Voxel-based models. PV-RCNN has integrated the advantages of both categories in one model. The structure of PV-RCNN is shown in Fig. \ref{fig:pv_rcnn}.

The model starts with voxelizing the input that is dividing it into smaller chunks of Pointclouds. The voxels are then fed into the 3D sparse convolutional encoder to extract multi-scale semantic features and generate 3D object proposal. Using a novel voxel set abstraction module, the extracted voxel-level features at multiple neural layers are summarised and aggregated to form a set of key points. Then the keypoint features are aggregated to the ROI-grid points to extract proposal specific features. Finally, a two-head fully connected network refines the proposed boxes and predicts a confidence score. 

\begin{figure}[!t]
     \centering
     \begin{subfigure}[b]{0.3\textwidth}
         \centering
         \includegraphics[width=\textwidth]{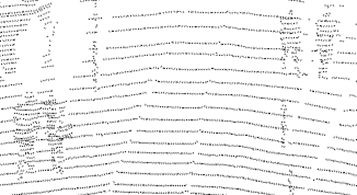}
         \caption{}
     \end{subfigure}
     \\
     \begin{subfigure}[b]{0.2\textwidth}
         \centering
         \includegraphics[width=\textwidth]{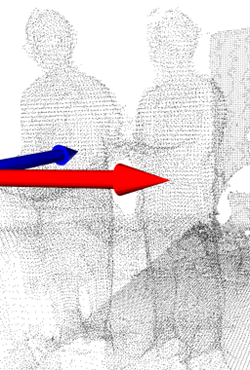}
         \caption{}
     \end{subfigure}
        \caption{Humans in (a) KITTI dataset  and (b) MAESTRO dataset}
        \label{fig:human_data}
\end{figure}

\subsubsection{Our modification}

As mentioned earlier, most 3D detectors (including PV-RCNN) are designed to work outdoors, specifically for autonomous driving applications/detection. Figure \ref{fig:human_data} shows a significant difference between the AV and MAESTRO Pointclouds. In MAESTRO, Humans are represented by a high number of points in a dense configuration (because they are close to the sensor) and the dimensions of humans are large compared to the size of the environment (operating room). In AV applications, humans are represented by a low number of relatively sparse pints and the dimensions of humans are relatively small compared to the environment (outdoor). Consequently, we had to modify many parameters to cope with the inherent differences in human representations and environments. 

\begin{itemize}
    \item [1.] Create a new python class and yaml file for a custom dataset (MAESTRO).
    \item [2.] The classes of interest, currently we are interested in detecting humans only.
    \item [3.] Point Cloud Range: The x, y, z ranges of interested. Those were in order of tens of meters in the original AV dataset, however, for MAESTRO, the range is limited to the size of the operating room which is in terms of a few meters.
    \item [4.] Voxel Size: Same rationale of the previous point, in the original AV datasets, the outdoor environment is large so the voxles are large. In the MAESTRO dataset, the environment is significantly smaller, so we minimised the voxel size.
    \item [5.] Anchor sizes: The starting point of the proposed bounding box. This can be decided based on our prior knowledge about the size of human, we changed to the average dimension of human in the MAESTRO Pointcloud.
    \item [6.] Other dimensions and parameters so that the dimensions of the tensor within the model are consistent (the dimension of the output of a certain layer equal to the dimension of the input of the next layer).
\end{itemize}

\section{Experimental Results}

\subsection{Endoscope Video Detection}
This subsection provides a detailed description of the structure of our novel endoscope dataset used for training and testing purposes. A description of the various experiments carried out and the resources used in training our models. We conducted an ablation study using the results of several experiments performed on our dataset to prove the feasibility and measure the performance of the modified and unmodified YOLOv5 models in recognizing surgical instruments from endoscopic videos. Finally, we show the results from training and testing our novel endoscope dataset with our benchmark models, and compare them with the top four models from our ablation study. 

\subsubsection{Dataset Structure}

\begin{figure}[!t]
\centering
\includegraphics[width= 0.5 \textwidth]{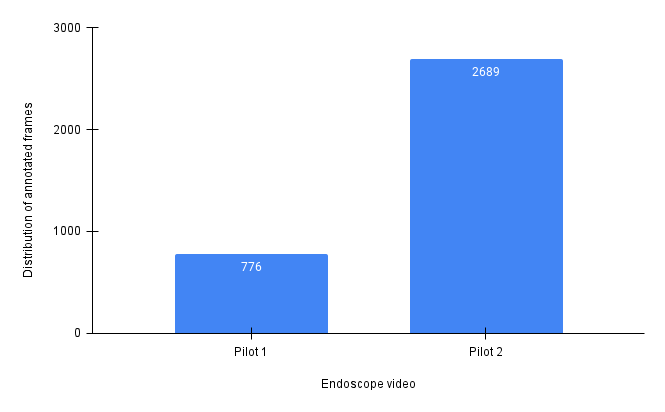}
\caption{\label{fig:chart_frames} A figure showing the distribution of the number of annotated frames in Pilot 1 and Pilot 2 endoscope videos.}
\end{figure}

\begin{figure}[!t]
\centering
\includegraphics[width= 0.5 \textwidth]{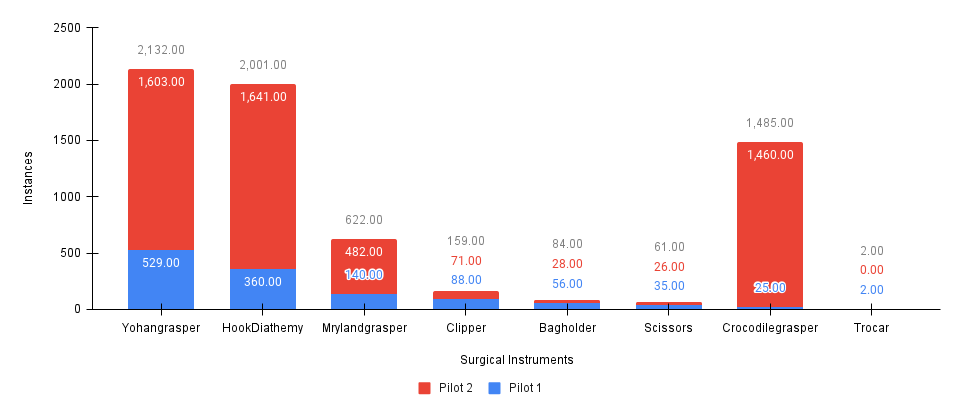}
\caption{\label{fig:chart_instances} Distribution of the samples per surgical instrument category in Pilot 1 and Pilot 2 endoscope videos.}
\end{figure}

Our novel endoscope dataset was extracted from two endoscope videos, referred to as Pilot 1 and Pilot 2. From Pilot 1, we annotated seven hundred and seventy-six image frames, and two thousand, six hundred and eighty-nine image frames were annotated from Pilot 2 (see Fig. \ref{fig:chart_frames}). Each image frame has a corresponding text file that contains the coordinates of the ground-truth annotation for each image. This file is saved in a different folder but in the same directory. To split our novel endoscope dataset into training and testing sets, we designed a Python script to randomly extract 70\% of Pilots 1 and 2 as the training sets, and the remaining 30\% of Pilots 1 and 2 were used as the test sets. This resulted in our training set consisting of 2425 image frames and their corresponding annotations, and our test set consisting of 1040 image frames and their corresponding annotations. Figure \ref{fig:chart_instances} displays the distribution of annotations per surgical instrument category for Pilot 1 and Pilot 2.

\begin{figure*}
     \centering
     \begin{subfigure}[b]{0.3\textwidth}
         \centering
         \includegraphics[width=\textwidth]{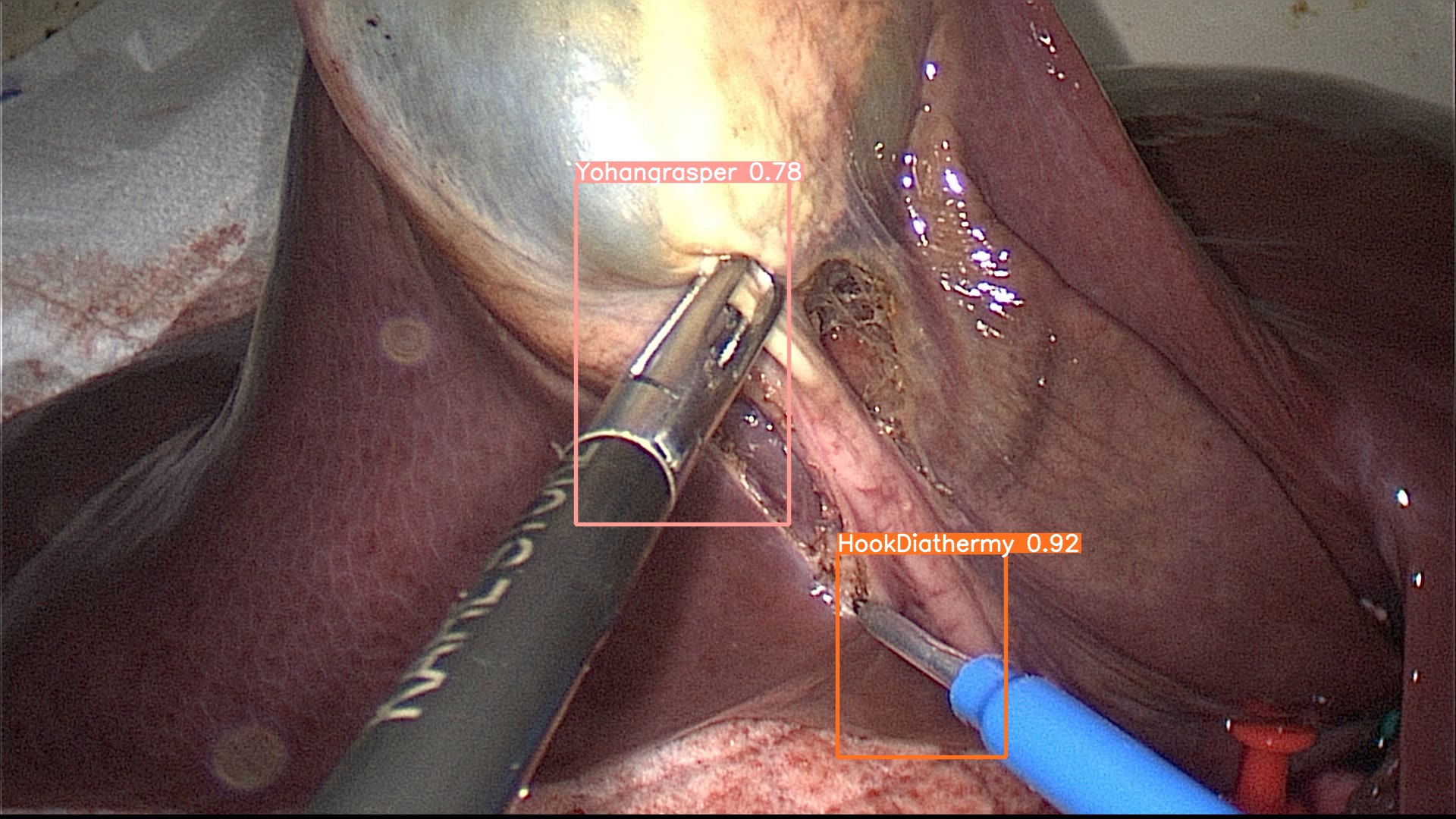}
     \end{subfigure}
     \begin{subfigure}[b]{0.3\textwidth}
         \centering
         \includegraphics[width=\textwidth]{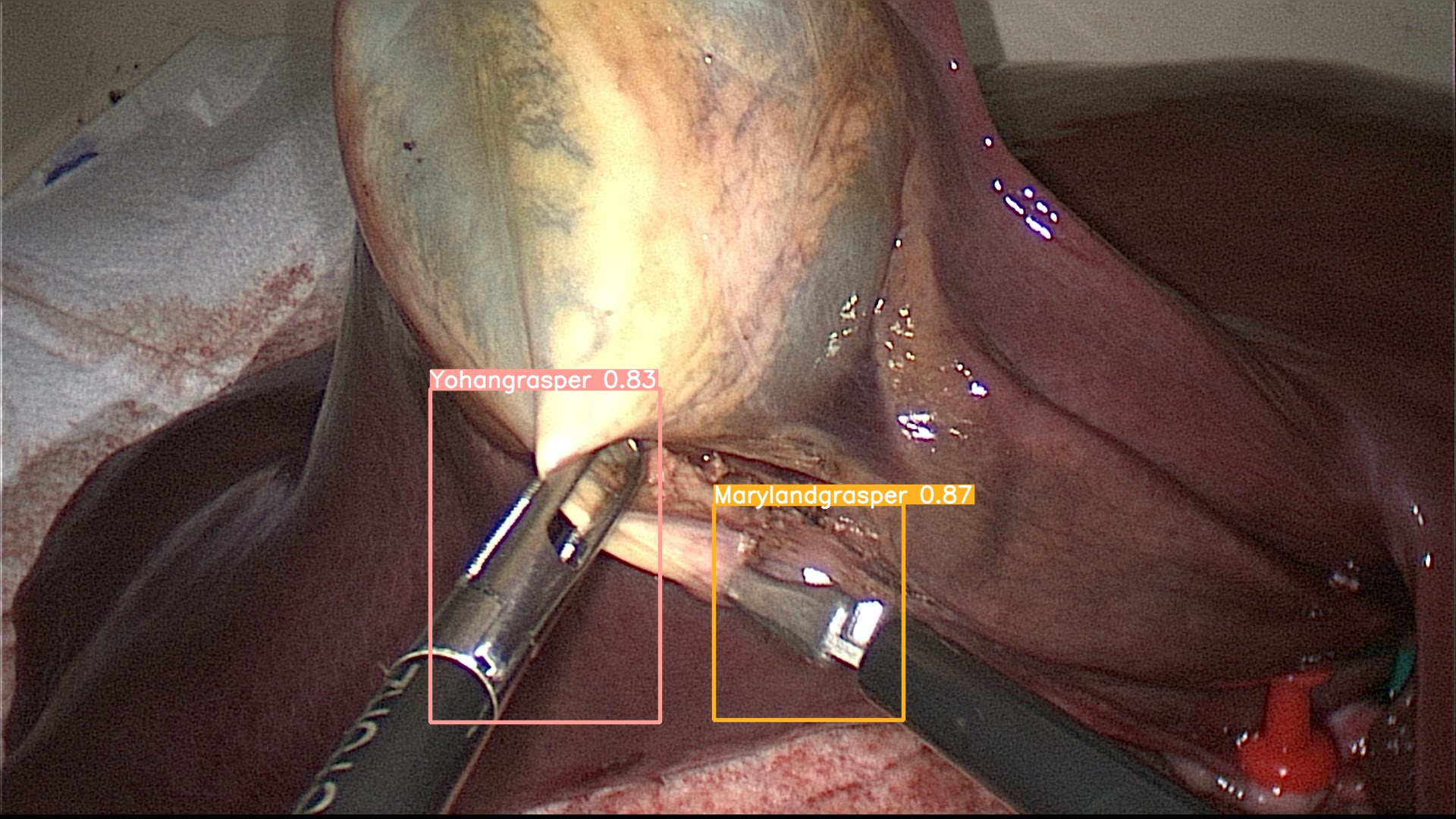}
     \end{subfigure}
     \begin{subfigure}[b]{0.3\textwidth}
         \centering
         \includegraphics[width=\textwidth]{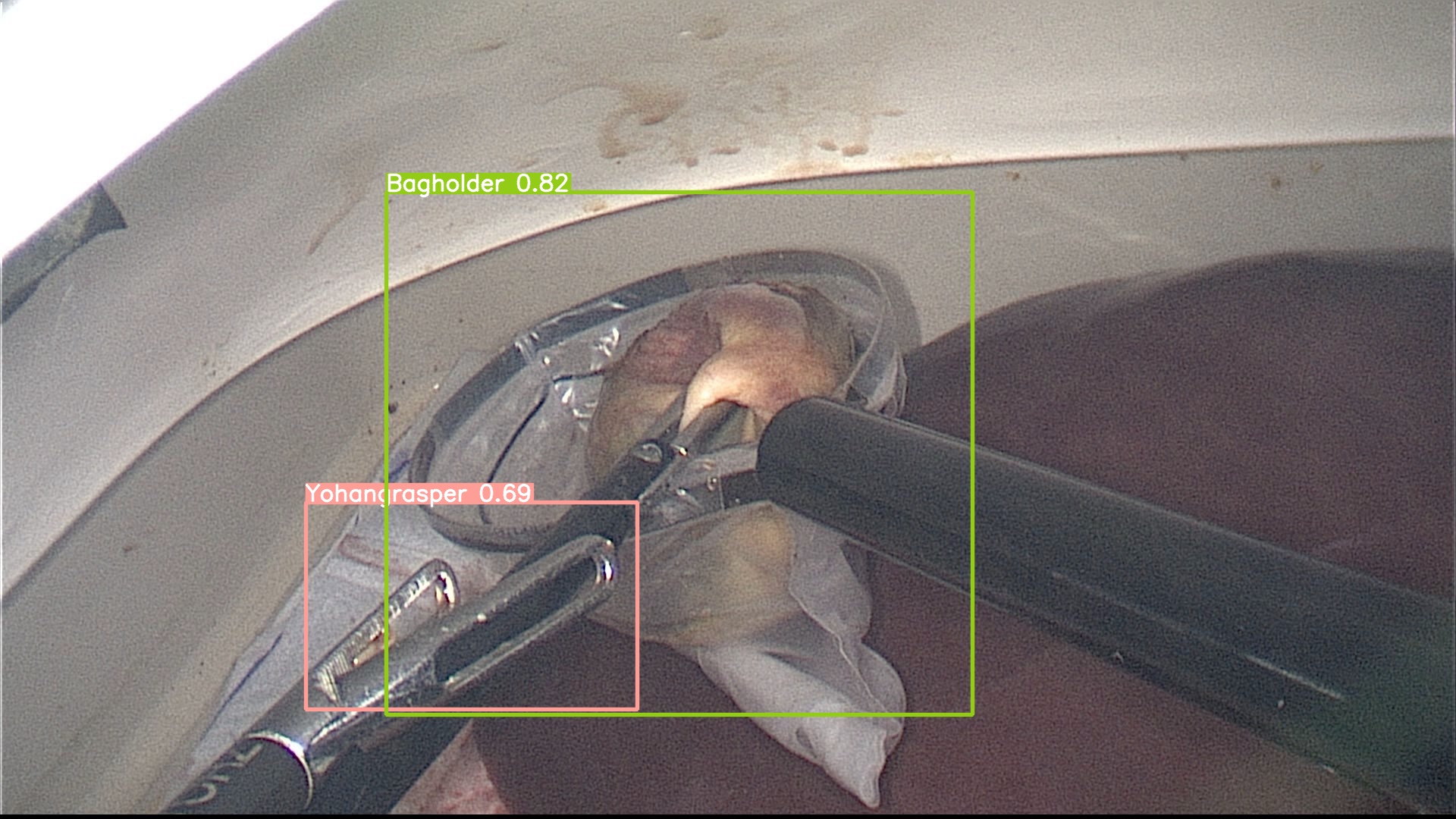}
     \end{subfigure}        
        \begin{subfigure}[b]{0.3\textwidth}
         \centering
         \includegraphics[width=\textwidth]{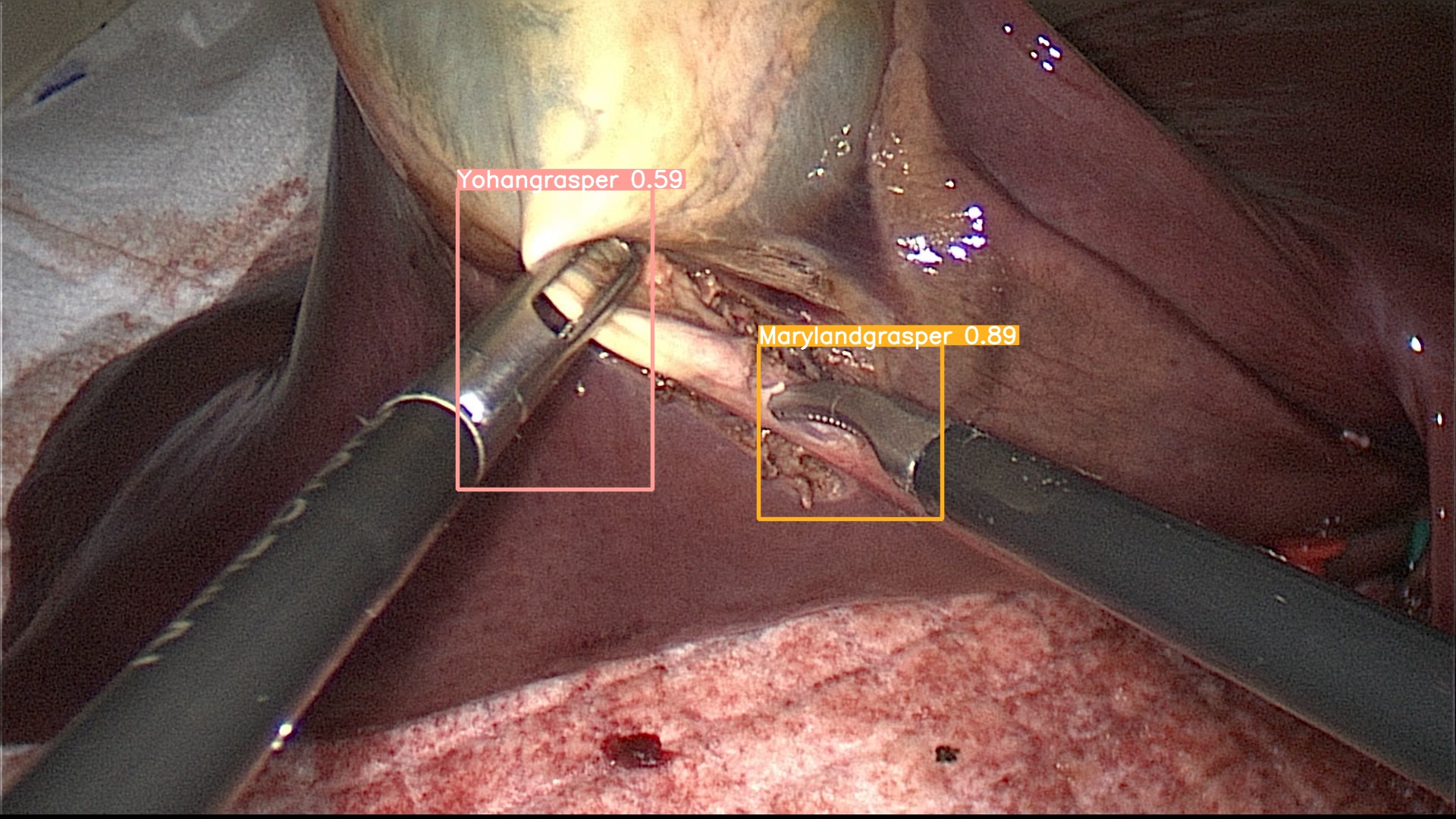}
     \end{subfigure}
     \begin{subfigure}[b]{0.3\textwidth}
         \centering
         \includegraphics[width=\textwidth]{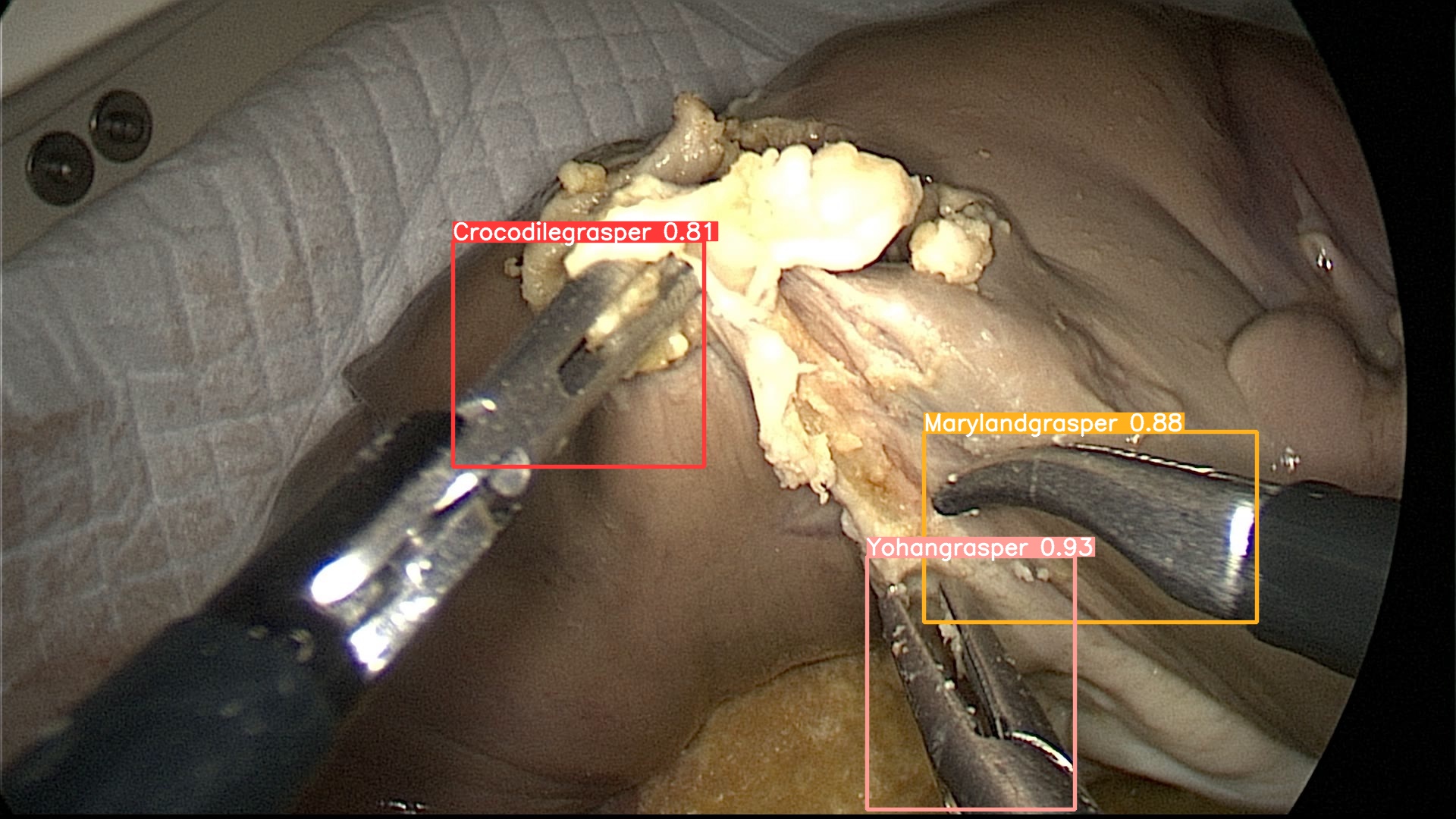}
     \end{subfigure}
     \begin{subfigure}[b]{0.3\textwidth}
         \centering
         \includegraphics[width=\textwidth]{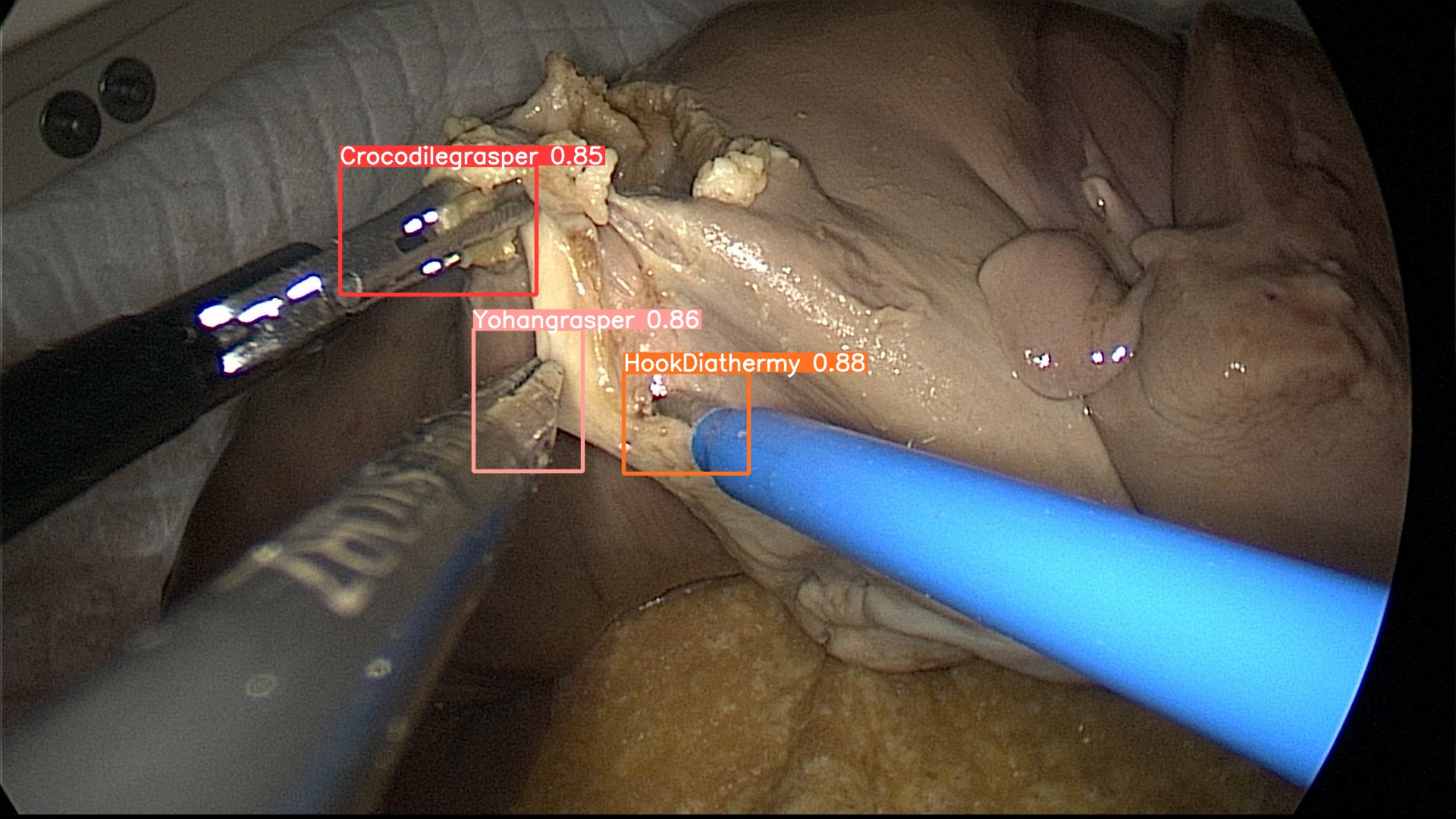}
     \end{subfigure}
        \caption{Samples of visualization results of surgical instrument detection using our model. Different colored bounding boxes represent different class categories in our dataset}
        \label{fig:figureview}
\end{figure*}

\subsubsection{Experimental Setup}
All experiments were trained and tested on the Visual Artificial Intelligence Laboratory beta-Mars server with an Ubuntu 20.04 operating system and PyTorch framework. The server has four Nvidia GTX 1080 GPUs, each of which has 12 GB VRAM. The training time for each experiment took an average of six hours to complete. In Fig. \ref{fig:figureview}, we demonstrate the visualization of surgical equipment detection using our model, where different colored bounding boxes correspond to various class categories in our dataset.

\subsubsection*{Hyperparameter Settings}

In deep learning problems, such as this one, hyperparameter setting is a fundamental task that is required to improve model performance. The grid search approach for selecting the best hyperparameter value for a model is often used during model training. However, a major disadvantage of this approach is that it requires a significant amount of runtime to train across all possible hyperparameter values, thereby increasing the computational complexity of the model. Owing to time constraints and limited access to computing resources, only a few hyperparameters were manually adjusted to select the best-performing model. Table \ref{tab:hy_para} displays the final values of the selected hyperparameters used for training all the models.

\begin{table}[t!]
\caption{\label{tab:hy_para} Hyperparameters for 2D Detection.}
\begin{center}
\begin{tabular}{c||c}
\hline
\textbf{Hyperparameter}  & \textbf{Value}  \\ \hline\hline
Learning rate start & 0.01 \\    \hline
Learning rate end & 0.1 \\    \hline
Flipud & 0.5 \\     \hline
Weight rate decay & 0.0005 \\     \hline
Momentum & 0.937 \\     \hline
Batch size & 16 \\     \hline
Epochs & 300 \\     \hline
Optimizer & Stochastic gradient descent \\     \hline
\end{tabular}
\end{center}
\end{table}

\subsubsection{Ablation Study}

In order to illustrate the efficacy of our proposed refinements to the YOLOv5 algorithm, we conducted an ablation study to demonstrate the effectiveness of each modification in an incremental manner using our novel endoscope dataset and YOLOv5l as our baseline model. F1 score, mAP@0.5, mAP@0.5:0.95, and inference are used as our evaluation metrics. The results of all experiments are summarized in Table \ref{tab:ablation}. 

\begin{table*}[!t]
    \centering
    \caption{An ablation study of model refinements on our novel endoscope dataset. We report the F1-Score, mAP@0.5 IOU, mAP@0.5:0.95 IOU, parameters, and the inference time.}
    \label{tab:ablation}    
    \begin{tabular}{c||c c c c c c c c c}
    \hline 
         & \textbf{Backbone} & \textbf{Neck} & \textbf{Anchors} & \textbf{Size} & \textbf{F1 score} & \textbf{\begin{tabular}[c]{@{}c@{}}mAP\\@0.5\end{tabular}} & \textbf{\begin{tabular}[c]{@{}c@{}}mAP\\@0.5:0.95\end{tabular}} & \textbf{Parameters} & \textbf{\begin{tabular}[c]{@{}c@{}}Inference \\ (ms)\end{tabular}}\\
         \hline \hline
       \textbf{A} & CSPDarknet53 & PANet & Predefined & 640 & 0.90 & 0.981 & 0.6 & 46145973 & 9.4 \\
       \hline
       \textbf{B} & CSPDarknet53 & PANet & 3 & 640 & 0.94 & 0.983 & 0.607 & 46145973 & 9.5 \\
       \hline
       \textbf{C} & CSPDarknet53 & PANet & 5 & 640 & 0.95 & 0.978 & 0.628 & 46192643 & 9.5 \\
       \hline
       \textbf{D} & CSPDarknet53 & BiFPN & 3 & 640 & 0.93 & 0.975 & 0.627 & 46408117 & 9.5 \\
       \hline
       \textbf{E} & CSPDarknet53 & BiFPN & 5 & 640 & 0.96 & 0.977 & 0.626 & 46454787 & 9.5 \\
       \hline
       \textbf{F} & CSPDarknet53 & FPN & 3 & 640 & 0.97 & 0.98 & 0.589 & 40703157 & 8.8 \\
       \hline
        \textbf{G} & CSPDarknet53 & FPN & 5 & 640 & 0.97 & 0.982 & 0.571 & 40749827 & 8.7 \\
       \hline
       \textbf{H} & Our backbone & PANet & 3 & 640 & 0.95 & 0.98 & 0.602 & 95906421 & 5.8 \\
       \hline
        \textbf{I} & Our backbone & PANet & 5 & 640 & 0.97 & 0.98 & 0.612 & 96019651 & 5.9 \\
       \hline
       \textbf{J} & Our backbone & FPN & 3 & 640 & 0.97 & 0.978 & 0.585 & 33039477 & 3.6 \\
       \hline
       \textbf{K} & Our backbone & FPN & 5 & 640 & 0.97 & 0.976 & 0.606 & 33086147 & 3.7 \\
       \hline \hline
    \end{tabular}
\end{table*}

\textbf{Model A:} To begin with, we build our baseline model using the original design of the YOLOv5 algorithm, which has predefined anchors. This resulted in mAP (98.1\%), F1-score (90\%), mAP@0.5 (98.1\%) and inference speed (9.8ms). The original YOLOv5 setting is described in Section \ref{overview}.

\textbf{Model B and C:} Model B is the first refinement with a positive effect on the YOLOV5 algorithm, and it involves replacing the predefined anchors with three auto-anchor generators. Auto-anchors is a technique introduced in YOLOv5 which helps the algorithm automatically learn the best anchor box for any given dataset. This boosted performance from 98.1\% mAP to 98.3\% mAP. Model C refinement is similar to model B, except that we used five auto-anchor generators. The model resulted in 97.8\% mAP, which is slightly lower than model B and our baseline model. Models B and C are slightly slower than model A. 

\textbf{Model D and E:} In these models, the refinements made were replacing the baseline YOLOv5 neck architecture, PANet with BiFPN, and experimenting with three and five auto-anchor generations. Both models resulted in mAP (97.5\% and 97.7\%) which are lower than our baseline model but achieve a higher F1-score when compared to our baseline model. Both models also achieved higher inference speeds when compared to our baseline model. 

\textbf{Model F and G:} The refinements made in these models are similar to those made in models D and E except that we used a different neck architecture, FPN. While model F achieved a slightly lower mAP (98.0\%) compared to our baseline model, model G achieved a slightly higher mAP value (98.2\%) than our baseline model, which makes it the second refinement with a positive effect on the YOLO algorithm. Both models F and G achieve higher inference speed than our baseline model. 

\textbf{Model H and I:} The refinements made in models H and I involved replacing the backbone of our baseline model with our light-weight backbone inspired by VGG and experimenting with three and five auto-anchor generation. Both models achieved the same mAP value 0.98\% which is slightly lower than our baseline model but had a higher inference speed and F1-score when compared to our baseline model.

\begin{table*}[!t]
    \centering
    \caption{The comparative results of the speed and accuracy of the top four model performances from our ablation studies with four other state-of-the-art object detection algorithms.}
    \label{tab:com_speed_acc}
    \begin{tabular}{c || c c c c c}
    \hline 
       \textbf{Model} & \textbf{F1 score} & \textbf{mAP@0.5} & \textbf{mAP@0.5:0.95} & \textbf{Parameters} & \textbf{Inference} \\
       \hline \hline 
       \textbf{YOLOv7} & 0.97 & 0.98 & 0.617 & 36519530 & 10.7 \\
       \hline
       \textbf{YOLOv3-SPP} & 0.93 & 0.983 & 0.513 & 62584213 & 12.3 \\
       \hline
       \textbf{Scaled-YOLOv4} & 0.81 & 0.836 & 0.469 & 52501333 & 11.8 \\
       \hline
       \textbf{YOLOR} & 0.86 & 0.832 & 0.47 & 36844024 & 23.5 \\
       \hline
       \textbf{B} & 0.94 & 0.983 & 0.607 & 46145973 & 9.5 \\
       \hline
       \textbf{G} & 0.97 & 0.982 & 0.571 & 40749827 & 8.7 \\
       \hline
       \textbf{A} & 0.90 & 0.981 & 0.6 & 46145973 & 9.4 \\
       \hline
       \textbf{I} & 0.97 & 0.98 & 0.612 & 96019651 & 5.9 \\
       \hline \hline
    \end{tabular}
\end{table*}

\begin{table*}[!t]
    \centering
    \caption{Results comparing mAP@0.5 value for each surgical instrument for the different model experiments conducted.}
    \label{tab:sur_inst}
    \begin{tabular}{ c || c  c  c c c c c c }
    \hline
    \textbf{Model} & \textbf{\begin{tabular}[c]{@{}c@{}}Crocodile \\ grasper\end{tabular}} & \textbf{\begin{tabular}[c]{@{}c@{}}Johan \\ grasper\end{tabular}} & \textbf{\begin{tabular}[c]{@{}c@{}}Hook\\diathermy \end{tabular}} & \textbf{\begin{tabular}[c]{@{}c@{}}Maryland \\ grasper\end{tabular}} & \textbf{Clipper} & \textbf{Scissors} & \textbf{\begin{tabular}[c]{@{}c@{}} Bag\\ holder \end{tabular}}& \textbf{Trocar} \\
    \hline \hline
    \textbf{A} & 0.978 & 0.956 & 0.978 & 0.986 & 0.995 & 0.995 & 0.961 & 0.995 \\
     \hline
    \textbf{B} & 0.977 & 0.965 & 0.981 & 0.995 & 0.955 & 0.995 & 0.962 & 0.995 \\
     \hline
    \textbf{C} & 0.971 & 0.943 & 0.979 & 0.984 & 0.995 & 0.995 & 0.958 & 0.995 \\
     \hline
    \textbf{D} & 0.968 & 0.947 & 0.978 & 0.978 & 0.995 & 0.995 & 0.946 & 0.995 \\
     \hline
    \textbf{E} & 0.975 & 0.94 & 0.978 & 0.984 & 0.995 & 0.955 & 0.955 & 0.955 \\
     \hline
    \textbf{F} & 0.97 & 0.952 & 0.982 & 0.984 & 0.995 & 0.995 & 0.965 & 0.995 \\
     \hline
    \textbf{G} & 0.977 & 0.96 & 0.986 & 0.986 & 0.995 & 0.995 & 0.962 & 0.995 \\
     \hline
    \textbf{H} & 0.974 & 0.949 & 0.976 & 0.99 & 0.995 & 0.995 & 0.965 & 0.995 \\
     \hline
    \textbf{I} & 0.961 & 0.956 & 0.981 & 0.991 & 0.995 & 0.995 & 0.964 & 0.995 \\
     \hline
    \textbf{J} & 0.951 & 0.95 & 0.986 & 0.995 & 0.995 & 0.995 & 0.959 & 0.995 \\
     \hline
    \textbf{K} & 0.95 & 0.953 & 0.979 & 0.984 & 0.995 & 0.995 & 0.96 & 0.995 \\
     \hline
    \textbf{YOLOv7} & 0.981 & 0.961 & 0.985 & 0.983 & 0.996 & 0.972 & 0.966 & 0.955 \\
     \hline
    \textbf{YOLOv3-SPP} & 0.983 & 0.96 & 0.977 & 0.995 & 0.995 & 0.995 & 0.964 & 0.995 \\
     \hline 
    \textbf{Scaled-YOLOv4} & 0.939 & 0.928 & 0.974 & 0.975 & 0.995 & 0.976 & 0.901 & 0 \\
     \hline
    \textbf{YOLOR} & 0.969 & 0.948 & 0.976 & 0.981 & 0.995 & 0.968 & 0.818 & 0 \\
    \hline \hline
    \end{tabular}
\end{table*}

\textbf{Model J and K:} The refinements made to models J and K are similar to models H and I except that we replace the PANet neck architecture in models H and I with an FPN neck architecture, experimenting with three and five auto-anchor generations. While both models J and K achieved mAP values (97.8\% and 97.6\%) slightly lower than our baseline model, both models were faster than our baseline model and also achieved an F1-score higher than that of our baseline model. 

\subsubsection{Comparison with Benchmark Models}
In this section, we compare the performances of the top four models from our ablation studies in Table \ref{tab:ablation} with four other state-of-the-art object detection models that can be used for surgical instrument detection tasks. Table \ref{tab:com_speed_acc} summarizes the comparison results, whereas Table \ref{tab:sur_inst} displays the performance of all the different experiments conducted in terms of their mean average precision (mAP) at a 50\% IOU threshold for each surgical instrument in our novel endoscope dataset. The results in Table \ref{tab:com_speed_acc} demonstrate that model B, which is the best-performing model from our ablation studies in Table \ref{tab:sur_inst}, outperformed all our benchmark models except for YOLOv3-SPP, which achieved equal model performance in terms of the mAP value.

\subsection{Rotated Surgical Tools Detection}
\begin{figure*}
     \centering
     \begin{subfigure}[b]{0.45\textwidth}
         \centering
         \includegraphics[width=\textwidth]{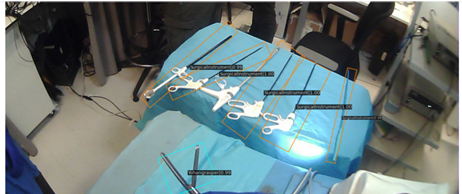}
     \end{subfigure}
     \begin{subfigure}[b]{0.4\textwidth}
         \centering
         \includegraphics[width=\textwidth]{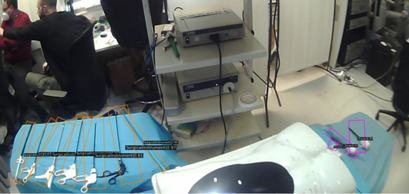}
     \end{subfigure}
     \begin{subfigure}[b]{0.45\textwidth}
         \centering
         \includegraphics[width=\textwidth]{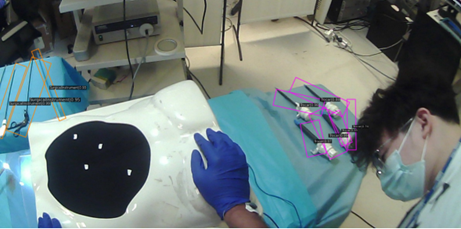}
     \end{subfigure}        
        \begin{subfigure}[b]{0.4\textwidth}
         \centering
         \includegraphics[width=\textwidth]{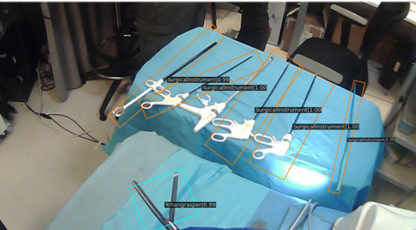}
     \end{subfigure}
        \caption{Visual results of surgical instruments and their tips detection using our rotated objection detection model.}
        \label{fig:rodet}
\end{figure*}

This section provides both quantitative and qualitative results of our rotated object detection model. Table \ref{tab:rodet} shows the performance of each class in terms of recall and average precision including the mAP for all classes. In addition, we visualized the performance of our rotated tool detector model in Fig. \ref{fig:rodet}, where we showed both the tips of the tools and entire tool.

\begin{table}[!t]
    \centering
    \caption{Results of our rotated object detection model for all classes with IoU threshold of 0.5.}
    \label{tab:rodet}
    \begin{tabular}{c || c c c c}
    \hline 
       \textbf{Classes} & \textbf{Gts} & \textbf{Dets} & \textbf{Recall} & \multicolumn{1}{c}{\textbf{\begin{tabular}[c]{@{}c@{}}Average \\ Precision\end{tabular}}}\\
       \hline \hline 
       \textbf{Clipper} & 7 & 35	& 0.286 & 	0.182  \\
       \hline
       \textbf{\begin{tabular}[c]{@{}c@{}}Crocodile \\ grasper\end{tabular}} & 11 & 21	& 0.727	& 0.655 \\
       \hline
       \textbf{\begin{tabular}[c]{@{}c@{}}Hook\\ diathermy\end{tabular} } & 6	& 36 &	1.00 &	0.939 \\
       \hline
       \textbf{\begin{tabular}[c]{@{}c@{}}Maryland\\ grasper\end{tabular}}  & 21 &	91 &	9.05 &	0.886 \\
       \hline
       \textbf{Scissors} & 18 &	58 &	0.944 &	0.868 \\
       \hline
       \textbf{\begin{tabular}[c]{@{}c@{}}Surgical\\ instrument\end{tabular}} & 1787 &	2478 &	0.628 &	0.471 \\
       \hline
       \textbf{Trocar} & 162 &	469	& 0.623	& 0.506 \\
       \hline
       \textbf{Johan grasper} & 31 &	81	& 0.935	& 0.894 \\ \hline
       \textbf{mAP} &   &   &   & \textbf{0.675} \\
       \hline \hline
    \end{tabular}
\end{table}

\subsection{Phase Segmentation}
\begin{figure}
    \centering
    \includegraphics[width=0.45 \textwidth]{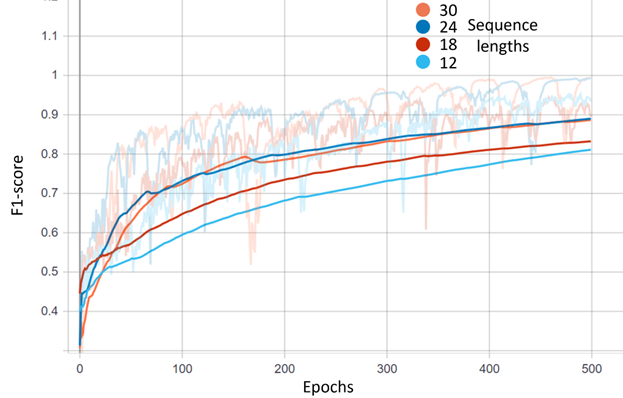}
    \caption{Fully-connected graph}
    \label{fig:phase_1}
\end{figure}

\begin{figure}
    \centering
    \includegraphics[width= 0.45 \textwidth]{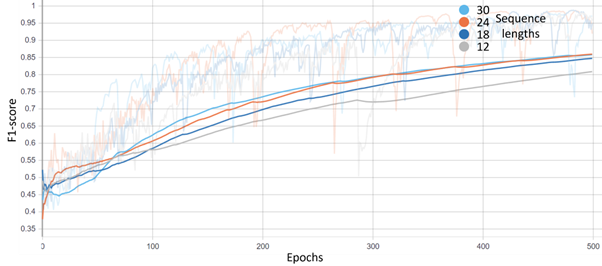}
    \caption{Scene graph}
    \label{fig:phase_2}
\end{figure}

\begin{figure}
    \centering
    \includegraphics[width= 0.45  \textwidth]{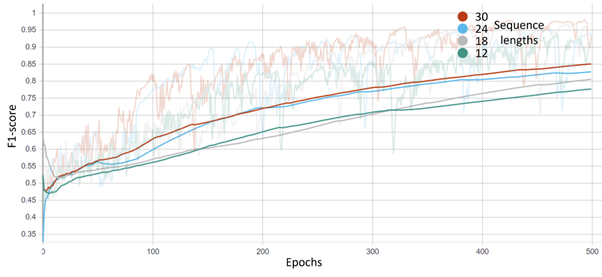}
    \caption{Scene with same agent labels graph}
    \label{fig:phase_3}
\end{figure}

This section presents the performance of the proposed phase segmentation model. We demonstrated the performance of all three combinations of models, including a fully connected graph, scene graph, and scene with the same label graph. For each graph, we selected four different sequence lengths as local graphs, i.e., 12, 18, 24, and 30 frames, as presented in Figures \ref{fig:phase_1}, \ref{fig:phase_2}, and \ref{fig:phase_3}.

\subsection{3D Detection}

\subsubsection{Experimental Setup}
We trained the model using a 12 GB VRAM Nvidia GTX 1080 GPU. The training time was 3 hours on average.

\subsubsection{Hyperparameter Settings}
Most deep-learning-based models are sensitive to hyperparameters, and the accuracy of the results depends heavily on the right choice of hyperparameter values. For training, we used the values presented in Table \ref{tab:hy_para_3d}.

\begin{table}[!t]
    \centering
    \caption{Hyperparameters for 3D Detection.}
    \begin{tabular}{c||c} \hline
        \textbf{Hyperparameter} & \textbf{Value} \\ \hline \hline
        Batch Size & 2 \\ \hline
        Number of Epochs & 80 \\ \hline
        Optimizer & ADAM \\ \hline
        Learning Rate & 0.01 \\ \hline
        Weight Decay & 0.01 \\ \hline
        Momentum  & 0.9 \\ \hline \hline
    \end{tabular}
    \label{tab:hy_para_3d}
\end{table}

\subsubsection{Preliminary Results}

\begin{figure}[!t]
     \centering
     \begin{subfigure}[b]{0.3\textwidth}
         \centering
         \includegraphics[width=\textwidth]{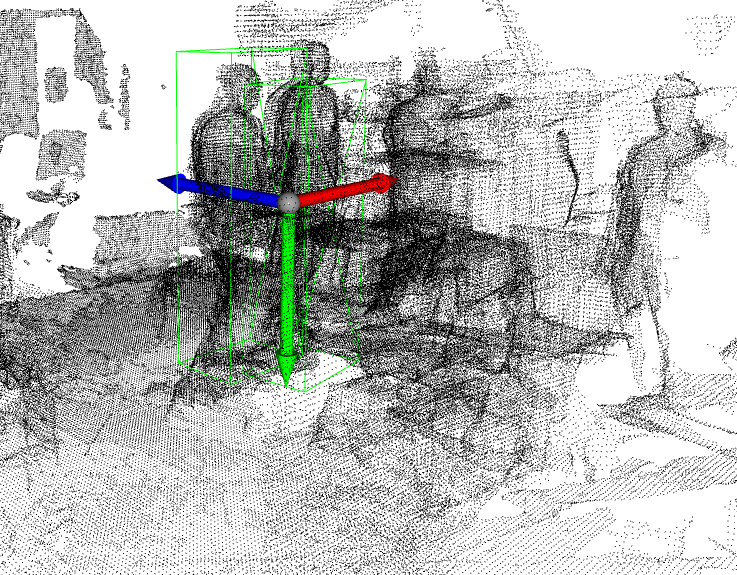}
         \caption{}
     \end{subfigure}
     \\
     \begin{subfigure}[b]{0.3\textwidth}
         \centering
         \includegraphics[width=\textwidth]{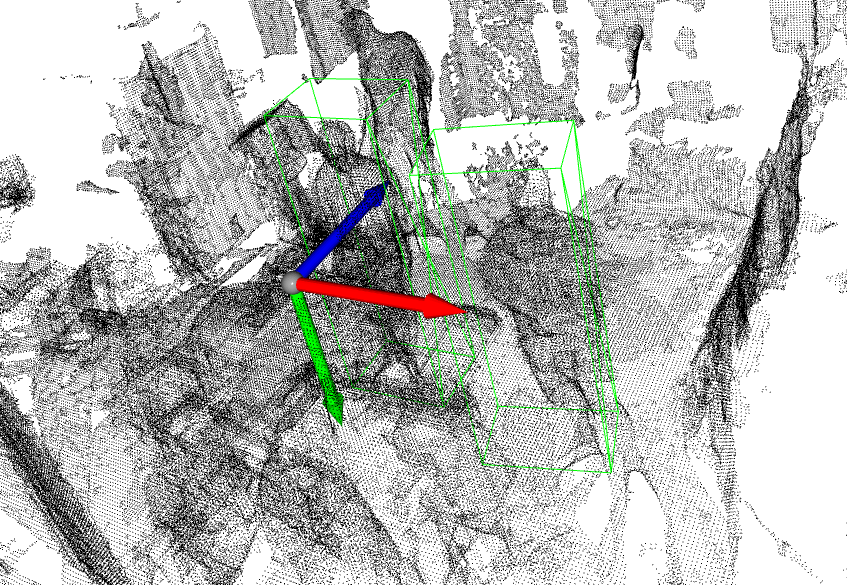}
         \caption{}
     \end{subfigure}
        \caption{Samples of 3D detection results}
        \label{fig:3d results}
\end{figure}

Fig. \ref{fig:3d results} shows a sample of the 3D detection results. We can see that the model detected two humans out of 4 in each image. This is due to the low number of training samples. Usually, 3D detection models are trained with ten times more samples than that used for the preliminary experiment. Therefore, the following training batch will contain at least five times the number used for this batch.

\section{Conclusion}
\label{con}
The detection and tracking of surgical instruments have become a crucial area of research because of the rising need for minimally invasive surgery in a variety of surgical procedures. Despite the recent development of numerous algorithms for this task, many issues remain to be resolved. Therefore, to recognize surgical tools in our novel endoscope dataset, we concentrated on constructing a deep-learning-based object detection approach in this study. This method entails four crucial steps: data processing of our unique endoscope dataset, YOLOv5 model improvement, performance evaluation through ablation tests, and benchmarking of the performance of the preceding step against four cutting-edge object detection models.

\section*{Acknowledgement}
Special thanks to the annotators: Dr Vivek Singh, Dr Ajmal Shahbaz, and Shireen Kudukkil Manchingal.

{\small
\bibliographystyle{ieee_fullname}
\bibliography{egbib}

\begin{thebibliography}{10}\itemsep=-1pt

\bibitem{ref27}
Alexey Bochkovskiy, Chien-Yao Wang, and Hong-Yuan~Mark Liao.
\newblock Yolov4: Optimal speed and accuracy of object detection.
\newblock {\em arXiv preprint arXiv:2004.10934}, 2020.

\bibitem{caesar2020nuscenes}
Holger Caesar, Varun Bankiti, Alex~H Lang, Sourabh Vora, Venice~Erin Liong,
  Qiang Xu, Anush Krishnan, Yu Pan, Giancarlo Baldan, and Oscar Beijbom.
\newblock nuscenes: A multimodal dataset for autonomous driving.
\newblock In {\em Proceedings of the IEEE/CVF conference on computer vision and
  pattern recognition}, pages 11621--11631, 2020.

\bibitem{geiger2012we}
Andreas Geiger, Philip Lenz, and Raquel Urtasun.
\newblock Are we ready for autonomous driving? the kitti vision benchmark
  suite.
\newblock In {\em 2012 IEEE conference on computer vision and pattern
  recognition}, pages 3354--3361. IEEE, 2012.

\bibitem{ref5}
Annetje~CP Gu{\'e}don, Linda~SGL Wauben, Anne~C van~der Eijk, Alex~SN Vernooij,
  Fr{\'e}d{\'e}rique~C Meeuwsen, Maarten van~der Elst, Vivian Hoeijmans, Jenny
  Dankelman, and John~J van~den Dobbelsteen.
\newblock Where are my instruments? hazards in delivery of surgical
  instruments.
\newblock {\em Surgical endoscopy}, 30(7):2728--2735, 2016.

\bibitem{ref87}
Iason Katsamenis, Eleni~Eirini Karolou, Agapi Davradou, Eftychios
  Protopapadakis, Anastasios Doulamis, Nikolaos Doulamis, and Dimitris
  Kalogeras.
\newblock Tracon: A novel dataset for real-time traffic cones detection using
  deep learning.
\newblock {\em arXiv preprint arXiv:2205.11830}, 2022.

\bibitem{ref77}
Kwang-Ju Kim, Pyong-Kun Kim, Yun-Su Chung, and Doo-Hyun Choi.
\newblock Performance enhancement of yolov3 by adding prediction layers with
  spatial pyramid pooling for vehicle detection.
\newblock In {\em 2018 15th IEEE International Conference on Advanced Video and
  Signal Based Surveillance (AVSS)}, pages 1--6. IEEE, 2018.

\bibitem{ref83}
Alex Krizhevsky, Ilya Sutskever, and Geoffrey~E Hinton.
\newblock Imagenet classification with deep convolutional neural networks.
\newblock {\em Advances in neural information processing systems}, 25, 2012.

\bibitem{ref68}
Tsung-Yi Lin, Piotr Doll{\'a}r, Ross Girshick, Kaiming He, Bharath Hariharan,
  and Serge Belongie.
\newblock Feature pyramid networks for object detection.
\newblock In {\em Proceedings of the IEEE conference on computer vision and
  pattern recognition}, pages 2117--2125, 2017.

\bibitem{ref85}
Tsung-Yi Lin, Michael Maire, Serge Belongie, James Hays, Pietro Perona, Deva
  Ramanan, Piotr Doll{\'a}r, and C~Lawrence Zitnick.
\newblock Microsoft coco: Common objects in context.
\newblock In {\em European conference on computer vision}, pages 740--755.
  Springer, 2014.

\bibitem{ref3}
David Orentlicher.
\newblock Medical malpractice: treating the causes instead of the symptoms.
\newblock {\em Medical Care}, 38(3):247--249, 2000.

\bibitem{ref4}
David Orentlicher.
\newblock Medical malpractice: treating the causes instead of the symptoms.
\newblock {\em Medical Care}, 38(3):247--249, 2000.

\bibitem{ref26}
Joseph Redmon and Ali Farhadi.
\newblock Yolov3: An incremental improvement.
\newblock {\em arXiv preprint arXiv:1804.02767}, 2018.

\bibitem{ref2}
WHO~Patient Safety, World~Health Organization, et~al.
\newblock {\em WHO guidelines for safe surgery 2009: safe surgery saves lives}.
\newblock Number WHO/IER/PSP/2008.08-1E. World Health Organization, 2009.

\bibitem{Shi2020}
Shaoshuai Shi, Chaoxu Guo, Li Jiang, Zhe Wang, Jianping Shi, Xiaogang Wang, and
  Hongsheng Li.
\newblock Pv-rcnn: Point-voxel feature set abstraction for 3d object detection.
\newblock {\em Proceedings of the IEEE Computer Society Conference on Computer
  Vision and Pattern Recognition}, pages 10526--10535, 2020.

\bibitem{ref64}
Karen Simonyan and Andrew Zisserman.
\newblock Very deep convolutional networks for large-scale image recognition.
\newblock {\em arXiv preprint arXiv:1409.1556}, 2014.

\bibitem{ref69}
Mingxing Tan, Ruoming Pang, and Quoc~V Le.
\newblock Efficientdet: Scalable and efficient object detection.
\newblock In {\em Proceedings of the IEEE/CVF conference on computer vision and
  pattern recognition}, pages 10781--10790, 2020.

\bibitem{ref82}
Do Thuan.
\newblock Evolution of yolo algorithm and yolov5: The state-of-the-art object
  detention algorithm.
\newblock 2021.

\bibitem{ref75}
Chien-Yao Wang, Alexey Bochkovskiy, and Hong-Yuan~Mark Liao.
\newblock Scaled-yolov4: Scaling cross stage partial network.
\newblock In {\em Proceedings of the IEEE/cvf conference on computer vision and
  pattern recognition}, pages 13029--13038, 2021.

\bibitem{ref29}
Chien-Yao Wang, Alexey Bochkovskiy, and Hong-Yuan~Mark Liao.
\newblock Yolov7: Trainable bag-of-freebies sets new state-of-the-art for
  real-time object detectors.
\newblock {\em arXiv preprint arXiv:2207.02696}, 2022.

\bibitem{ref84}
Chien-Yao Wang, I-Hau Yeh, and Hong-Yuan~Mark Liao.
\newblock You only learn one representation: Unified network for multiple
  tasks.
\newblock {\em arXiv preprint arXiv:2105.04206}, 2021.

\bibitem{ref1}
World Health~Organization (WHO) et~al.
\newblock Patient safety fact file: patient safety and risk management service
  delivery and safety.
\newblock {\em WHO: Geneva, Switzerland}, page~14, 2019.

\bibitem{xie2021oriented}
Xingxing Xie, Gong Cheng, Jiabao Wang, Xiwen Yao, and Junwei Han.
\newblock Oriented r-cnn for object detection.
\newblock In {\em Proceedings of the IEEE/CVF International Conference on
  Computer Vision}, pages 3520--3529, 2021.

\end{thebibliography}
}

\end{document}